\documentclass{article}

\usepackage{PRIMEarxiv}

\usepackage[utf8]{inputenc} 
\usepackage[T1]{fontenc}    
\usepackage{hyperref}       
\usepackage{url}            
\usepackage{booktabs}       
\usepackage{amsfonts}       
\usepackage{nicefrac}       
\usepackage{microtype}      
\usepackage{lipsum}
\usepackage{fancyhdr}       
\usepackage{graphicx}       
\graphicspath{{media/}}     
\usepackage{authblk}
\usepackage{amsmath} 
\usepackage{tabularx}
\title{Virtual iEEG from Scalp EEG: Charting the Landscape of Source Imaging, Intracranial Inference and Reconstruction}

\author[1]{Dongyi He}
\author[2]{Xiangkai Wang}
\author[3]{Hongjie Yan}
\author[4]{Luping Song}
\author[1,*]{Wai Ting Siok}
\author[1,*]{Nizhuan Wang}

\affil[1]{Department of Language Science and Technology, The Hong Kong Polytechnic University, Hung Hom, Hong Kong SAR, China}
\affil[2]{School of Artificial Intelligence, Chongqing University of Technology, Chongqing, China}
\affil[3]{Affiliated Lianyungang Hospital of Xuzhou Medical University, Lianyungang, China}
\affil[4]{School of Artificial Intelligence, Southern University of Science and Technology Hospital, Shenzhen, China}
\affil[*]{Correspondence: wai-ting.siok@polyu.edu.hk (W.T.S.); wangnizhuan1120@gmail.com (N.W.)}

\begin{document}

\maketitle

\begin{abstract}
Intracranial electroencephalography (iEEG) provides temporally precise and spatially specific access to neural activity from focal and deep brain regions, but its invasiveness and restricted anatomical coverage limit routine use. These constraints have motivated scalp-to-intracranial inference, termed virtual iEEG when model outputs carry iEEG-defined event, feature, representation, or contact-level waveform semantics. This review presents a target-centred framework distinguishing event inference, feature translation, and waveform reconstruction, while separating predictability from observability, identifiability, fidelity, and utility. Evidence is evaluated according to cohort independence, anatomical and spectral coverage, train--test separation, and target-patient adaptation. Current studies support inference of selected intracranial events, low-frequency components, and task-related representations, but not unique recovery of arbitrary contact-level activity. Stronger validation requires appropriate controls, source-imaging baselines, uncertainty assessment, and incremental-utility testing. Future progress depends on independent paired datasets and prospective evidence that virtual iEEG adds value beyond scalp EEG and EEG source imaging.
\end{abstract}


\section{Introduction}

\begin{figure*}[!t]
    \centering
    \includegraphics[width=\textwidth]{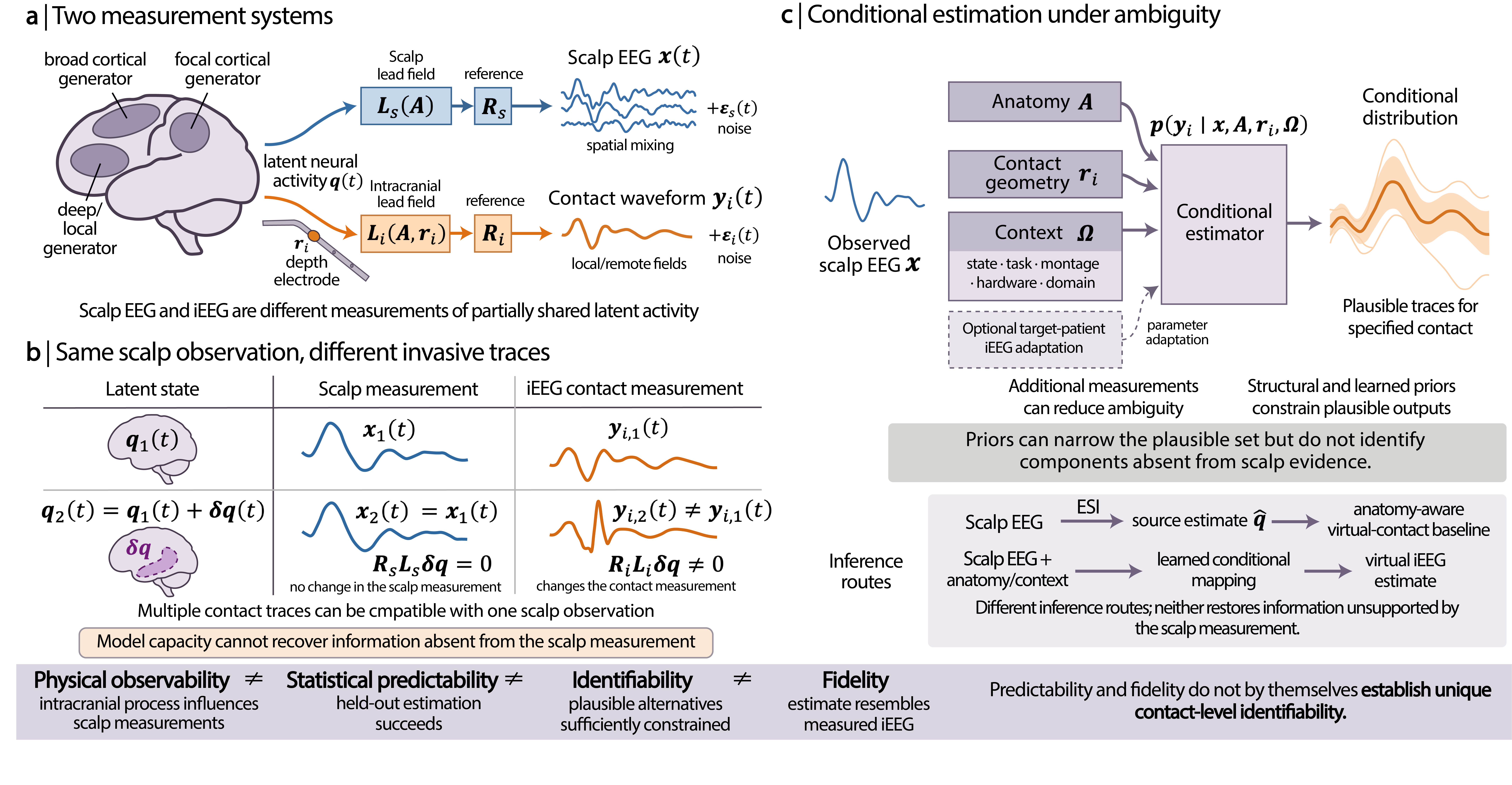}
    \caption{\textbf{Target-centred framework for scalp-to-intracranial EEG inference.}
    \textbf{(a)} Available information comprises scalp EEG \(\boldsymbol{x}\), anatomy \(\boldsymbol{A}\), target-contact geometry \(\boldsymbol{r}_i\), deployment context \(\boldsymbol{\mit\Omega}\), and, where permitted, target-patient iEEG adaptation.
    \textbf{(b)} Prediction targets comprise event inference, feature translation, and waveform reconstruction, with waveform outputs represented as point estimates or conditional distributions.
    \textbf{(c)} Evidential claims are distinguished by observability, predictability, identifiability, fidelity, and utility.}
    \label{fig:framework}
\end{figure*}

Intracranial electroencephalography (iEEG) records electrical activity directly from within the brain, providing greater spatial specificity and bandwidth than scalp EEG. Stereoelectroencephalography (SEEG) depth electrodes can access mesial temporal, insular, and other deep structures, whereas electrocorticography (ECoG) grids and strips sample cortical fields with high spatial and temporal resolution~\cite{lachaux2012high,cardinale2013stereoelectroencephalography,parvizi2018promises}. These recordings have transformed epilepsy surgery as well as human cognitive and brain--computer interface (BCI) research~\cite{rosenow2001presurgical,kwan2011drug,engel2012early,Anumanchipalli2019,Willett2021,Moses2021,zhang2026imagined,he2026toward}. Their use is constrained by neurosurgical implantation, availability only in clinically selected patients, and sampling restricted to regions targeted by a diagnostic hypothesis. Even direct intracranial recordings therefore provide an anatomically sparse and clinically conditioned view of brain activity.

These limitations motivate inference of intracranial activity from non-invasive recordings. Scalp EEG is attractive because it is safe, inexpensive, portable, and temporally precise~\cite{berger1929elektroenkephalogramm,schomer2018niedermeyer,nunez1982electric,ebersole2014current}, but neural currents are spatially mixed and attenuated before being sampled as reference-dependent boundary potentials by a sparse sensor array~\cite{malmivuo1995bioelectromagnetism,nunez1982electric,einevoll2013modelling,jiang2026comprehensive,li2025tale}. Scalp EEG is therefore neither a direct measurement of a local generator nor uniformly informative about intracranial activity~\cite{li2025tale}. Magnetoencephalography (MEG)-based virtual iEEG demonstrates that selected intracranial features can be estimated non-invasively~\cite{Cao2022,Afnan2026}, but conventional MEG remains less accessible than scalp EEG.

Existing work relevant to scalp-to-intracranial inference spans two related but distinct methodological lines. EEG source imaging estimates neural generators or source-space activity from scalp recordings using explicit forward models and inverse constraints, and therefore provides an anatomy-aware reference framework rather than virtual iEEG itself~\cite{brodbeck2011electroencephalographic,Michel2019}. In parallel, virtual iEEG methods target iEEG-defined outputs, including hidden-event inference~\cite{Spyrou2016IED,lam2016widespread,Lam2017}, intracranial feature or representation inference~\cite{Antoniades2018,zhang2025cross,Yang2026CORTEG}, and explicit contact-level waveform reconstruction~\cite{Hu2022E2SGAN,He2026NeuroFlowNet,Pham2026CAST,Dong2026Bridge}. Source imaging may localise generators without reproducing SEEG morphology~\cite{brodbeck2011electroencephalographic,Michel2019}; event detectors may identify iEEG-labelled states without recovering contact waveforms~\cite{Spyrou2016IED,lam2016widespread,Lam2017}; generative models may produce plausible iEEG-like signals only weakly constrained by the concurrent scalp segment~\cite{Hu2022E2SGAN,He2026NeuroFlowNet,Dong2026Bridge}; and cross-modal representation methods may improve or transfer invasive representations without necessarily predicting an iEEG waveform at deployment~\cite{zhang2025cross,Yang2026CORTEG}. The evidential strength therefore depends on the prediction target and information flow rather than the model architecture alone.

The available evidence supports a bounded account of scalp-to-intracranial inference. Scalp EEG contains useful information about selected intracranial events, low-frequency components, and task-related representations, but successful prediction does not imply unique recovery of arbitrary contact-level activity. The relevant question is which intracranial targets are predictable from scalp EEG and under what conditions. Recoverability depends on anatomy, frequency, state, population, acquisition context, and adaptation of the target-patient, while residual uncertainty may remain irreducible. \textit{Scalp-to-intracranial reconstruction (STIR)} denotes this broader task family, whereas \textit{virtual iEEG} denotes model outputs assigned intracranial semantics; neither implies equivalence to direct intracranial measurement.

Consequently, this review progresses from the definition of target and biophysical limits of recoverability to EEG source imaging (ESI) as an anatomical reference, empirical evidence for hidden-event inference and intracranial representations, and explicit contact-level waveform reconstruction. It then examines probabilistic reconstruction, predictive uncertainty, and generalisation across patients and deployment settings, before defining claim-specific validation requirements and translational pathways. The resulting synthesis provides a target-centred framework linking event inference, feature translation, and waveform reconstruction to observability, predictability, identifiability, fidelity, and utility; audits the evidence with respect to cohort independence, anatomical and spectral scope, train--test separation, and target-patient adaptation; and identifies the conditions required to demonstrate scalp-conditioned reconstruction and incremental value beyond scalp EEG and ESI. The contributions of this review can be summarised as follows:
\begin{itemize}
    \item A target-centred conceptual framework that distinguishes STIR and virtual iEEG from ESI, organises scalp-to-intracranial inference into event inference, feature translation, and waveform reconstruction, and separates observability, predictability, identifiability, fidelity, and utility.
    \item A structured evidence and validation framework that evaluates current studies in terms of cohort independence, anatomical and spectral coverage, train--test separation, and target-patient iEEG adaptation, while defining claim-specific requirements for deterministic and probabilistic reconstruction, including conditioning controls, uncertainty calibration, and matched scalp-EEG and ESI baselines.
    \item A translational and future-research roadmap that identifies the requirements for establishing incremental utility beyond scalp EEG and ESI, and outlines key priorities including independent paired datasets, anatomy- and frequency-stratified evaluation, calibrated uncertainty and abstention, target-iEEG-free cross-patient and cross-site generalisation, prospective validation, and systematic characterisation of the boundaries of recoverability.
\end{itemize}

\begin{table*}[!t]
\caption{Comparison of EEG source imaging and scalp-to-intracranial reconstruction across targets, modelling assumptions, validation criteria, and evidential limits}
\centering
\small
\begin{tabularx}{\linewidth}{p{0.08\linewidth} p{0.22\linewidth} p{0.23\linewidth} X}
\toprule
\textbf{Dimension} & \textbf{Shared basis} & \textbf{EEG source imaging (ESI)} & \textbf{Scalp-to-intracranial reconstruction (STIR)} \\
\midrule
\textbf{Input} & Volume-conducted, reference-dependent, sparsely sampled scalp potentials~\cite{Michel2019} & Boundary measurements for inverse estimation & Conditioning inputs for virtual-contact estimates or representations \\
\midrule
\textbf{Biophysics} & Neural currents; head anatomy; scalp lead field & Explicit forward model; regularised or prior-based inverse~\cite{Michel2019} & Paired-data conditional mapping; implicit biophysical relationships; geometric or physical constraints~\cite{Antoniades2018,Hu2022E2SGAN} \\
\midrule
\textbf{Target} & Latent neural information beyond raw scalp traces & Source location; current-density map; source-space time series & iEEG-defined event; time--frequency feature; embedding; contact waveform \\
\midrule
\textbf{Predictive form} & Deterministic or probabilistic inference & Regularised or prior-constrained source estimate & Point estimate or conditional distribution \\
\midrule
\textbf{Anatomy} & Electrode position; head model; cortical geometry; source depth & Head model; source-space definition & Contact semantics; representation localisation; anatomy-stratified prediction \\
\midrule
\textbf{Data} & Anatomical and acquisition metadata & Scalp EEG and head model; optional paired-iEEG validation & Paired or aligned scalp--iEEG; supervised target learning \\
\midrule
\textbf{Validation} & Anatomy-aware evaluation; non-invasive comparators & Localisation error, spatial accuracy, and stimulation validation~\cite{Unnwongse2023}; resection or clinical-outcome concordance~\cite{brodbeck2011electroencephalographic} & Waveform and spectral fidelity or event performance~\cite{Antoniades2018,Hu2022E2SGAN}; for probabilistic outputs, calibration and proper scoring~\cite{Gneiting2007}; downstream utility where claimed \\
\midrule
\textbf{Failure mode} & Reference or conductivity error; sensor sparsity; overinterpretation & Non-uniqueness; spatial blur or mislocalisation; dependence on priors, regularisation, and forward-model accuracy & Coverage-specific matching and arbitrary-position generation remain unresolved~\cite{Hu2022E2SGAN}; extratemporal transfer remains untested and conditional uncertainty was not evaluated~\cite{Antoniades2018} \\
\midrule
\textbf{Clinical meaning} & Decision support; hidden-generator hypotheses & Generator localisation; epileptogenic-region hypotheses & Virtual-contact or decoder-representation hypotheses \\
\bottomrule
\end{tabularx}
\label{tab:esi_vs_virtual_ieeg}
\end{table*}

\section{Prediction targets and evidential scope}

Terminology in this literature is heterogeneous, reflecting differences in prediction target, modelling assumptions, and intended use. Terms including ``reconstruction'', ``translation'', ``enhancement'', ``source imaging'', ``virtual electrode'', and ``decoding'' have been applied to partially overlapping tasks. The corresponding estimands nevertheless entail different comparators, metrics, and evidential claims. Table~\ref{tab:esi_vs_virtual_ieeg} identifies the shared basis and distinct scope of ESI and STIR.

\subsection{Target semantics and forms of inference}

As summarised in Fig.~\ref{fig:framework}(a), scalp-to-intracranial inference is conditioned on the available information, including scalp EEG \(\boldsymbol{x}\), anatomical information \(\boldsymbol{A}\), target contact geometry \(\boldsymbol{r}_i\), and acquisition or deployment context \(\boldsymbol{\mit\Omega}\). Optional target-patient iEEG adaptation represents an additional information source when paired invasive recordings are available. As shown in Fig.~\ref{fig:framework}(b), scalp-to-intracranial inference comprises three distinct target classes: event inference, feature translation, and waveform reconstruction. \textit{Scalp-to-intracranial reconstruction} (STIR) denotes methods that use scalp EEG to estimate an iEEG-defined event, feature, waveform, or representation. Predictions for each target may take the form of a point estimate or a conditional distribution. The more specific terms \textit{EEG-to-SEEG translation} and \textit{EEG-to-ECoG translation} identify the intended intracranial modality. \textit{Virtual iEEG} denotes outputs assigned intracranial semantics. By contrast, \textit{EEG source imaging} (ESI) refers to inverse estimation of generators or source-space activity under a forward head model and regularisation~\cite{Hamalainen1994,PascualMarqui1994,Dale2000,Michel2004,Grech2008,Michel2019}.

Three analytical dimensions characterise these outputs. Target semantics distinguish \textit{event inference}, \textit{feature translation}, and \textit{waveform reconstruction}, which concern event occurrence or timing, intracranial representations, and time-resolved traces at specified contacts, respectively~\cite{lam2016widespread,Lam2017,AbouJaoude2022,Antoniades2018,Abdi2021CPD,Hu2022E2SGAN,He2026NeuroFlowNet,Pham2026CAST}. Predictive form distinguishes deterministic from probabilistic estimation, with either form applicable to any target; \textit{probabilistic reconstruction} denotes estimation of a conditional distribution. The third dimension concerns intended use and the associated utility claim, with \textit{incremental utility} referring to improvement in a clinical or BCI decision beyond scalp EEG, ESI, and other available information. These dimensions describe different aspects of scalp-to-intracranial inference rather than successive levels of methodological maturity. An event detector may provide clinical utility even when contact-level waveform recovery is not feasible~\cite{lam2016widespread,Lam2017,AbouJaoude2022}, whereas a realistic generated waveform provides limited evidence of reconstruction if its timing and morphology are not constrained by the concurrent scalp input~\cite{Hu2022E2SGAN,Abdi2025VAE,He2026NeuroFlowNet}. The strength of the resulting claim therefore depends on the target, predictive form, and intended use rather than on model architecture alone.

\subsection{Predictability and identifiability}

As distinguished in Fig.~\ref{fig:framework}(c), prediction performance and physical identifiability describe different properties even after the target has been specified. For a contact-indexed target, the variables \(\boldsymbol{x}\), \(\boldsymbol{Y}_i\), \(\boldsymbol{A}\), and \(\boldsymbol{r}_i\) denote the scalp EEG, the specified iEEG target at contact \(i\), anatomy, and target-contact geometry, respectively. The variable \(\boldsymbol{\mit\Omega}\) denotes acquisition and deployment context, including hardware, montage, state, task, and the patient or population domain. A supervised model estimates a property of \(\boldsymbol{p}(\boldsymbol{Y}_i\mid \boldsymbol{x},\boldsymbol{A},\boldsymbol{r}_i,\boldsymbol{\mit\Omega})\) which does not invert a one-to-one physical transformation. Under squared error, the optimal deterministic predictor is the conditional mean \(\boldsymbol{m}_i(\boldsymbol{x},\boldsymbol{A},\boldsymbol{r}_i,\boldsymbol{\mit\Omega})=\mathbb{E}[\boldsymbol{Y}_i\mid \boldsymbol{x},\boldsymbol{A},\boldsymbol{r}_i,\boldsymbol{\mit\Omega}],\) and its risk decomposes as
\begin{equation}
\begin{aligned}
&\mathbb{E}\bigl\|
    \boldsymbol{Y}_i
    - f_\theta(\boldsymbol{x},\boldsymbol{A},\boldsymbol{r}_i,\boldsymbol{\mit\Omega})
\bigr\|_2^2
\\
&\qquad=
\mathbb{E}\!\left[
    \operatorname{tr}\operatorname{Var}
    \bigl(\boldsymbol{Y}_i
    \mid \boldsymbol{x},\boldsymbol{A},\boldsymbol{r}_i,\boldsymbol{\mit\Omega}\bigr)
\right]
\\
&\qquad\quad+
\mathbb{E}\bigl\|
    \boldsymbol{m}_i(\boldsymbol{x},\boldsymbol{A},\boldsymbol{r}_i,\boldsymbol{\mit\Omega})
    - f_\theta(\boldsymbol{x},\boldsymbol{A},\boldsymbol{r}_i,\boldsymbol{\mit\Omega})
\bigr\|_2^2 .
\end{aligned}
\label{eq:conditional_risk}
\end{equation}
The first term represents irreducible conditional variance. It includes local activity that never reaches the scalp, ambiguity among source configurations, unmodelled anatomy or contact geometry, reference effects, stochastic influences, and task variables unavailable to the model. The second term represents reducible estimation error within the model. Additional data or improved architectures can reduce this estimation error but cannot remove the irreducible variance. Informative measurements may reduce ambiguity, whereas structural priors constrain plausible estimates without identifying activity absent from scalp data.

Reported waveform similarity consequently provides evidence about predictability under the evaluated distribution, rather than unique identification of the realised contact trace. A mean-squared-error model can obtain a favourable score by reproducing common low-frequency structure while suppressing rare transients. A generative model can likewise sample realistic transients from an iEEG prior when the scalp observation does not determine which transient occurred. Evidence for identifiability therefore depends on sensitivity to the conditioning input, comparison with anatomically matched alternatives, and uncertainty that expands where the conditional distribution is broad.

Five related concepts capture the resulting evidential distinctions. \textit{Observability} concerns whether an intracranial process influences scalp measurements, whereas \textit{predictability} concerns whether that influence supports held-out estimation. \textit{Identifiability} concerns whether the target is sufficiently constrained to distinguish plausible alternatives, whereas \textit{fidelity} quantifies resemblance to measured iEEG. \textit{Utility} concerns whether use of the estimate improves a decision. Evidence for one property does not necessarily establish the others, and each is associated with a different evaluation objective.

\subsection{Virtual contacts as measurement-level targets}

The phrase \textit{virtual electrode} has different meanings across this literature. In ESI, it usually denotes a source-space time series at an anatomical coordinate. In reconstruction, it may denote a synthetic contact trace, a contact-conditioned latent variable, or a decoder input. Neither output is equivalent to a physical intracranial measurement. A real contact records a mixture shaped by its geometry, local and remote fields, tissue, amplifier, reference, and network state. Across four implanted patients, independent-component analysis showed that the dominant component explained a widely varying fraction of each iEEG channel's variance, with the pooled mean below 50\%~\cite{whitmer2010utility}. Although the cohort was small, the result is consistent with iEEG channels being mixtures rather than elemental neural sources.

Reproducing an iEEG channel therefore estimates a measurement-level target rather than an elemental neural quantity. Two approaches provide complementary information even when virtual iEEG is computationally downstream of ESI. ESI supplies an explicit anatomical reference frame and a physically interpretable source-space baseline. Paired learning instead estimates contact-specific measurement relationships that are difficult to encode analytically. If ESI localises an event but a learned model does not reproduce its morphology, the local detail may be less observable, more difficult to estimate, or both. Matched baselines provide a more direct test of these explanations than comparisons between architectures. Conversely, performance under anatomical mismatch or temporal shifting may indicate reliance on coverage-specific or marginal-prior shortcuts. Negative controls of this kind delineate the boundary of the measurement problem.

\section{Biophysical determinants of observability}

\begin{figure*}[!t]
    \centering
    \includegraphics[width=\textwidth]{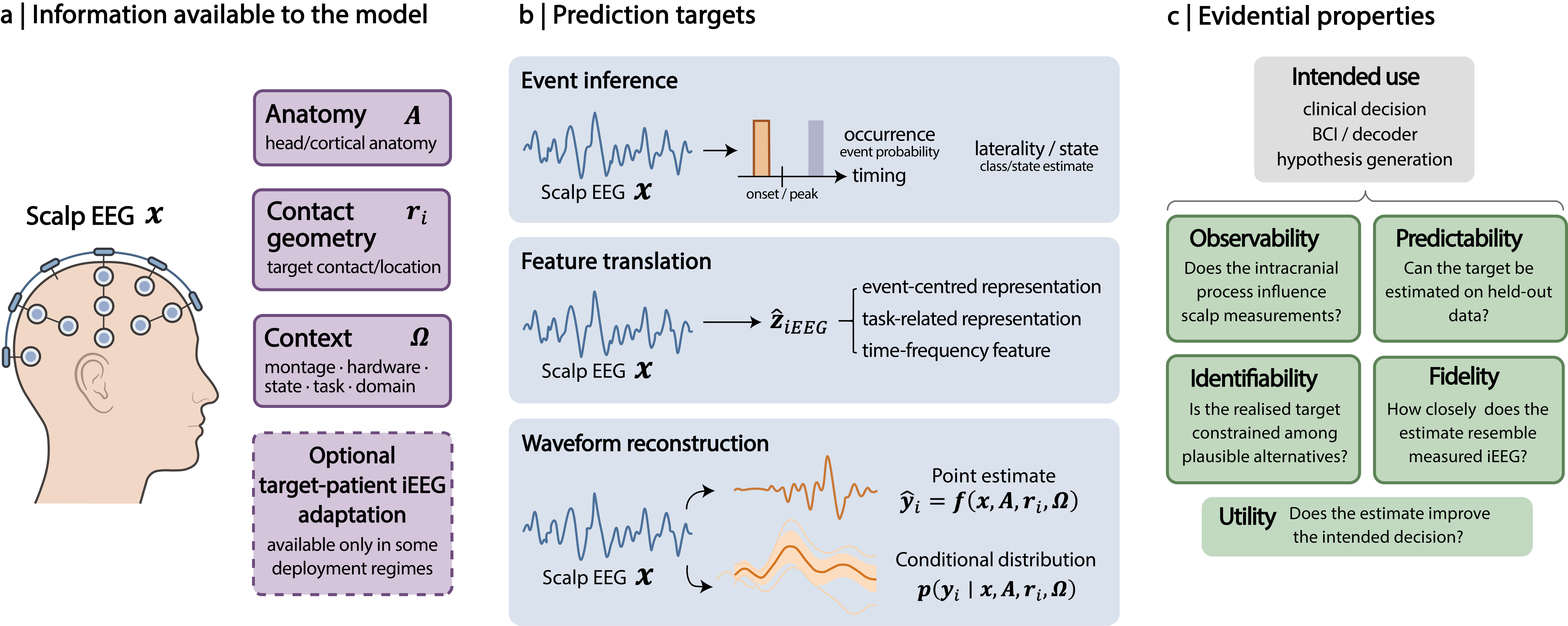}
    \caption{\textbf{Measurement model and identifiability limits in scalp-to-intracranial inference.}
    \textbf{(a)} Scalp EEG \(\boldsymbol{x}(t)\) and contact waveform \(y_i(t)\) are distinct, reference-dependent measurements of partially shared neural activity \(\boldsymbol{q}(t)\).
    \textbf{(b)} A scalp-null perturbation can alter the intracranial waveform without changing the scalp observation, limiting contact-level identifiability.
    \textbf{(b)} Reconstruction is therefore formulated as conditional estimation \(\boldsymbol{p}(\boldsymbol{y}_i\mid \boldsymbol{x},\boldsymbol{A},\boldsymbol{r}_i,\boldsymbol{\mit\Omega})\), constrained by anatomy, context, and optional adaptation; priors cannot recover information absent from scalp measurements.}
    \label{fig:measurement_identifiability}
\end{figure*}

Figure~\ref{fig:measurement_identifiability} summarises the measurement-level basis of scalp-to-intracranial inference, the resulting limit on contact-level identifiability, and the corresponding conditional-estimation formulation. As illustrated in Fig.~\ref{fig:measurement_identifiability}(a), scalp EEG and iEEG are distinct measurements of partially shared latent neural activity. The variables \(\boldsymbol{q}(t)\), \(\boldsymbol{A}\), and \(\boldsymbol{r}_i\) denote latent state, anatomy, and the geometry of target contact \(i\), respectively; \(\boldsymbol{r}_s\) and \(\boldsymbol{r}_i\) denote the scalp and intracranial reference operators. The two measurements are
\begin{equation}
\begin{aligned}
\boldsymbol{x}(t)&=\boldsymbol{r}_s\boldsymbol{L}_s(\boldsymbol{A})\boldsymbol{q}(t)+\boldsymbol{\varepsilon}_s(t),\\
\boldsymbol{y}_i(t)&=\boldsymbol{r}_i\boldsymbol{L}_i(\boldsymbol{A},\boldsymbol{r}_i)\boldsymbol{q}(t)+\boldsymbol{\varepsilon}_i(t).
\end{aligned}
\label{eq:two_measurements}
\end{equation}
The null-space case in Fig.~\ref{fig:measurement_identifiability}(b) makes the identifiability limit explicit. Considering the noiseless measurement components, or equivalently holding the measurement-noise realisations fixed, if a perturbation \(\delta\boldsymbol{q}\) satisfies \(\boldsymbol{r}_s\boldsymbol{L}_s(\boldsymbol{A})\delta\boldsymbol{q}=0\) but \(\boldsymbol{r}_i\boldsymbol{L}_i(\boldsymbol{A},\boldsymbol{r}_i)\delta\boldsymbol{q}\neq0\), the scalp measurement can remain unchanged while the target-contact waveform changes. Information carried only by such a component is therefore not identifiable from scalp EEG alone. This ambiguity motivates the conditional formulation in Fig.~\ref{fig:measurement_identifiability}(c): scalp-to-contact inference concerns \(\boldsymbol{p}(\boldsymbol{y}_i\mid \boldsymbol{x},\boldsymbol{A},\boldsymbol{r}_i,\boldsymbol{\mit\Omega})\), where \(\boldsymbol{\mit\Omega}\) denotes acquisition and deployment context, including hardware, montage, state, task, and the patient or population domain. Additional measurements may reduce ambiguity, whereas structural and learned priors constrain plausible outputs without identifying components absent from scalp evidence.

This formulation identifies the physical basis of both measurable and missing components. Large, synchronised postsynaptic currents in aligned cortical populations generate fields detectable at a distance. Different source configurations can nevertheless cancel or produce nearly identical scalp maps~\cite{nunez1982electric,buzsaki2004neuronal,einevoll2013modelling}. Physical observability is consequently not a fixed property of a region. It varies with generator extent, orientation, synchrony, conductivity, sensor geometry, montage, and reference. Empirical detectability also depends on noise, repetition, task locking, preprocessing, and the criterion used to classify an event as visible~\cite{Vorwerk2014,Strobbe2014,AkalinAcar2016}.

Frequency-dependent scalp detectability does not arise from simple high-frequency filtering by the skull. Under the quasi-static approximation used by conventional EEG forward models, the lead field is approximately frequency independent. High-gamma activity and high-frequency oscillations are generally harder to observe because their generators tend to be focal and weakly synchronous. Such activity is also readily cancelled at a distance and contaminated by scalp muscle or instrumentation noise. Slow rhythms, by comparison, more often reflect spatially extended and coherent neural populations. Frequency therefore covaries with generator scale and signal-to-noise ratio rather than independently guaranteeing visibility~\cite{nunez1982electric,einevoll2013modelling}. Evidence from low-frequency activity does not establish recoverability of contact-specific high-frequency biomarkers.

Reference choice adds a further source of ambiguity. Scalp EEG and iEEG are both reference-dependent measurements, although reference effects differ across modalities. A common average scalp reference, a bipolar SEEG montage, and a depth-contact reference can emphasise different spatial scales. A learned mapping may consequently capture reference conventions as well as physiology. Explicit reporting identifies the reference convention, while reasonable reference perturbations test whether performance persists beyond it~\cite{Pigorini2024,Wang2019EECoG}.

\begin{table*}[!t]
\caption{Biophysical and empirical determinants of scalp observability, with conditions favouring or limiting intracranial-signal detectability}
\centering
\small
\begin{tabularx}{\linewidth}{p{0.11\linewidth}X X X}
\toprule
\textbf{Determinant} & \textbf{Mechanism} & \textbf{Favourable conditions} & \textbf{Limiting conditions} \\
\midrule
Source extent and synchrony & Net scalp amplitude depends on effective generator extent and field summation; synchrony increases summation but is not a binary prerequisite & Extended recruitment; coherent or otherwise constructively summating generators; high event-to-background SNR & Small effective extent; destructive cancellation; low SNR; empirical area thresholds are not universal~\cite{Tao2005,vonEllenrieder2014,vonEllenrieder2016} \\
\midrule
Depth and orientation & Depth, cortical geometry, source orientation, and sensor coverage shape lead-field amplitude and SNR; none alone fixes visibility & Strong scalp projection; prominent deep rhythms under intracranial-informed analysis & Weak projection; low-SNR pure mesial-temporal single events; current subcortical evidence is limited to prominent 8--10-Hz rhythms~\cite{Ahlfors2010,koessler2015catching,Seeber2019} \\
\midrule
Frequency content & Quasi-static propagation is approximately frequency independent; detectability instead depends on generator extent and frequency-specific background SNR & Sufficient extent and SNR; repeated delta signatures or prominent 8--10-Hz target rhythms & Focal, low-amplitude high-frequency generators require favourable SNR; low-frequency evidence does not establish contact-specific HFO recovery~\cite{vonEllenrieder2014,DeStefano2022,Seeber2019} \\
\midrule
Background state & Neural state changes the recorded activity and can alter scalp--intracranial comparison performance & Pre-specified and matched state; state-stratified evaluation & Pooling wake, NREM sleep, or anaesthesia without stratification~\cite{Wang2019EECoG,DeStefano2022} \\
\midrule
Event structure and analysis & Known event timing permits aggregation; averaging tests a repeatable scalp contribution, not single-trial recovery & Repeated, consistently timed events; event-locked or trigger-averaged analysis & Irregularly timed events; low-SNR single-trial targets; extrapolating averaged results to single events~\cite{koessler2015catching} \\
\midrule
Reference and montage & Re-referencing is spatial processing and can alter amplitude, polarity, spectral power, and coherence & Explicitly reported scalp and iEEG montages; sensitivity analysis across reasonable references & Unreported or mismatched references; treating montage-specific results as reference-invariant~\cite{Yao2007Reference,Zaveri2006} \\
\midrule
Anatomical registration & Head segmentation, tissue conductivities, and scalp-electrode co-registration determine forward solutions and localisation error & Individual MRI or electrode-position-warped template; digitised and co-registered sensors; explicit tissue model & Electrode misregistration; inaccurate conductivity; omitted conductive compartments; template-only geometry~\cite{AkalinAcar2016,Vorwerk2014} \\
\bottomrule
\end{tabularx}
\label{tab:observability_determinants}
\end{table*}

Simultaneous scalp--intracranial recordings illustrate how source geometry, signal-to-noise ratio, montage, and analysis criteria affect the label ``scalp-negative''. Tao et al. reviewed approximately 600 intracranial spikes from 16 patients and found recognisable scalp correlates for 90\% of spikes engaging more than \(10\,\mathrm{cm}^2\) of cortex~\cite{Tao2005}. In comparison, correlates were recognisable for only 10\% below that area and for none below \(6\,\mathrm{cm}^2\). These values are cohort- and criterion-specific rather than universal physical thresholds. Synchrony, orientation, background, montage, and visual sensitivity also affect recognisability~\cite{vonEllenrieder2014,vonEllenrieder2016}. Koessler et al. analysed 4,048 SEEG spikes from seven highly selected patients. The 1,949 pure mesial-temporal events had a mean scalp amplitude of only \(7.1\,\mu\mathrm{V}\), with a single-event signal-to-noise ratio of \(-2.1\) dB. Nevertheless, all nine pure mesial networks produced a significant contribution after SEEG-triggered averaging~\cite{koessler2015catching}. The aggregate contribution shows that single-event visual absence does not imply absence of a repeated-event scalp signal. Single-trial recovery was not evaluated; the significant scalp contribution emerged only after SEEG-triggered averaging.

Deep-source studies further separate conditional detectability from unconstrained recovery. Seeber et al. enrolled four patients with deep-brain stimulation (DBS) and excluded one without a clear intracranial alpha peak~\cite{Seeber2019}. Their analysis was restricted to 8--10-Hz closed-eye alpha in the remaining three patients. Source maxima were 14.8--23.5 mm from the implanted contacts, and the DBS time series defined the covariate used to identify the scalp source. The study demonstrates conditional detectability of a prominent low-frequency rhythm under target-informed analysis. Blind recovery of arbitrary subcortical activity was not evaluated. A hippocampal event may similarly be predictable through a widespread network response even when the local contact field lacks a unique scalp projection. Direct field observability, network-mediated predictability, and contact identifiability therefore represent separate evidential properties.

Across these recordings, demonstrated recoverability varies along several interacting dimensions rather than a universal depth or frequency hierarchy. Repeated, task-locked, low-frequency, spatially coherent events may be predictable even at depth, whereas a brief superficial focal event may not be. Global correlation can conceal poor performance in clinically important deep contacts and the selection of favourable contacts. Table~\ref{tab:observability_determinants} organises these dimensions across structure, depth, frequency, event type, state, scalp distance, and reference convention. Reporting the fraction of contacts or events for which prediction was attempted further distinguishes general coverage from selected performance.

\section{EEG source imaging as an anatomical reference frame}

The anatomical dependence of scalp observability makes ESI a physically grounded reference frame for scalp-to-intracranial inference. Within the framework of Fig.~\ref{fig:framework}(c), ESI primarily addresses source localisation, observability, and anatomical plausibility, whereas virtual iEEG additionally requires evidence of target-specific predictability, fidelity, and, where claimed, downstream utility. ESI estimates source-space activity from scalp potentials through an explicit forward model and an inverse solution constrained by smoothness, sparsity, anatomy, or statistical priors~\cite{Hamalainen1994,Dale2000,PascualMarqui2002,Michel2004,Grech2008}. In epilepsy, high-density EEG combined with individual anatomy can provide clinically informative localisation of irritative or seizure-related activity~\cite{brodbeck2011electroencephalographic,Lantz2003,Plummer2019,Sharma2019}. This established anatomical role makes ESI a relevant comparator for claims that learned scalp-to-intracranial models recover information beyond conventional non-invasive inference.

The inverse problem remains non-unique because many intracranial source configurations can generate similar scalp topographies. Regularisation therefore selects one estimate from a family of scalp-compatible solutions, with the resulting source time series depending on the head model, sensor geometry, noise assumptions, reference, and prior. With a fixed inverse operator \(\boldsymbol{W}_\lambda\), an anatomy-aware virtual-contact baseline for target contact \(i\) can be expressed as
\begin{equation}
\boldsymbol{\hat{y}}_{ESI,i}(t)
=\boldsymbol{r}_i\boldsymbol{L}_i(\boldsymbol{A},\boldsymbol{r}_i)\hat{\boldsymbol{q}}(t),
\qquad
\hat{\boldsymbol{q}}(t)=\boldsymbol{W}_\lambda\boldsymbol{x}(t),
\label{eq:esi_virtual_contact}
\end{equation}
where \(\boldsymbol{x}(t)\) denotes the scalp EEG measurement, \(\hat{\boldsymbol{q}}(t)\) the regularised source estimate, and \(\boldsymbol{r}_i\boldsymbol{L}_i(\boldsymbol{A},\boldsymbol{r}_i)\) the intracranial lead-field and reference operation associated with the specified contact. The resulting signal is a forward projection of an estimated source configuration into an intracranial montage rather than a reconstruction of the physical contact measurement. Components eliminated by the scalp measurement null space or suppressed by the inverse prior cannot be restored by this projection. ESI-derived virtual contacts therefore provide an explicit anatomy-aware baseline without implying contact-level identifiability.

Validation of ESI spans stimulation, simultaneous spontaneous intracranial recordings, simulations, and clinical outcomes, but these reference standards support different claims. Intracranial stimulation provides a known perturbation site under high-signal conditions, whereas spontaneous SEEG provides physiological timing but sparse and clinically determined anatomical sampling. Surgical outcome offers clinically relevant patient-level evidence but constitutes a spatially coarse endpoint influenced by multimodal presurgical information. Simulations provide complete source truth only within the assumed conductivity, geometry, source, and noise models. Localisation errors obtained under these regimes are therefore complementary rather than directly interchangeable.

Controlled stimulation illustrates the gap between favourable-condition localisation and prospective performance. In Localize-MI, seven patients underwent 61 stimulation sessions, with 4,800 inverse configurations evaluated per session~\cite{Mikulan2020}. Mean localisation error reached \(5.39\pm2.61\) mm after selecting the best configuration for each session, whereas errors across the full configuration space extended to approximately 50 mm. Unnwongse et al. analysed 341 accepted stimulation potentials from nine patients using 37-channel scalp EEG~\cite{Unnwongse2023}. Localisation errors ranged from 10.3 to 26.0 mm across conductivity and depth settings and increased from \(14.4\pm7.0\) mm for sources within 10 mm of the inner skull to \(23.4\pm6.4\) mm for sources deeper than 40 mm under the standard conductivity setting. These results establish that ESI can localise selected intracranial perturbations under controlled conditions while also demonstrating substantial sensitivity to depth and modelling choices.

Spontaneous-event studies provide a less controlled but physiologically closer test. Zauli et al. selected eight patients from a cohort of 50 and averaged SEEG-defined interictal epileptiform discharges (IEDs) before evaluating 27 dSPM/eLORETA configurations per patient~\cite{Zauli2024}. Mean localisation error was \(19.20\pm9.74\) mm after patient-wise selection of the best configuration and \(35.56\pm18.01\) mm across all configurations, with a maximum of 79.18 mm. Only \(9.4\pm7.7\%\) of the SEEG events were recognisable in a conventional 30-channel scalp montage. The approximately 2-cm best-case result therefore depended on invasive event timing, averaging, and configuration selection rather than prospective single-event recovery. The comparison reinforces the distinction between demonstrating a reproducible scalp contribution and identifying the corresponding intracranial measurement.

Clinical and simultaneous-recording studies extend this evidence to patient-level and ictal-network localisation. Brodbeck et al. reported 84.1\% sensitivity and 87.5\% specificity for high-resolution ESI using individual MRI, although these estimates were obtained from 52 of 152 operated patients~\cite{brodbeck2011electroencephalographic}. In the full cohort, low-resolution EEG with a template head model yielded substantially lower sensitivity and specificity. FAST-IRES reported localisation errors of \(18.1\pm14.08\) mm for IEDs and \(6.0\pm5.8\) mm for seizures in 36 operated patients, while direct iEEG localisation comparisons were available in only 16~\cite{Sohrabpour2020}. In simultaneous ictal recordings, an ICA--dipole analysis retained eight of 14 patients with usable scalp--invasive correspondence; error approached 10 mm for favourable superficial sources but averaged approximately 47 mm across retained patients and increased with depth~\cite{Barborica2021}. Collectively, these studies support anatomy- and network-level localisation under selected conditions, but they do not establish recovery of continuous contact-specific waveforms.

These validation regimes position ESI as both an anatomical reference and a diagnostic baseline for virtual iEEG. For waveform reconstruction, comparison with an ESI-derived source or virtual-contact estimate tests whether learned mappings provide predictive information beyond an explicit biophysical inverse. For event inference, source-space features can determine whether apparent gains arise from regional isolation already achievable non-invasively. For representation or decoder-transfer claims, sensor-space and source-space comparators evaluated under identical data partitions can distinguish cross-modal benefit from improved non-invasive feature extraction. Anatomy-stratified comparisons are additionally informative because reconstruction confidence would be expected to deteriorate where the lead field and empirical ESI evidence indicate weak observability.

ESI also provides a structured route for incorporating anatomy into learned scalp-to-intracranial models. Data-driven mappings can capture empirical scalp--intracranial relationships that are difficult to specify analytically, whereas head models, contact geometry, and cortical distance can constrain those mappings to anatomically plausible regimes. The complementary strengths do not remove the underlying measurement limits. Conductivity assumptions, skull defects, co-registration error, electrode density, source depth, reference choice, and regularisation can materially alter ESI estimates~\cite{Vorwerk2014,Strobbe2014,AkalinAcar2016,Unnwongse2023}, while source-space connectivity remains sensitive to leakage and mixing~\cite{Schoffelen2009,Palva2018,Wang2019EECoG}. Deep-learning source-imaging methods inherit these dependencies and introduce additional sensitivity to the simulations, anatomical priors, and training distributions used for optimisation~\cite{Cui2019,pantazis2021meg,hecker2021convdip,Yu2024STSIN,Wu2025DMLN,sun2023deep}.

Distribution shift therefore limits both source imaging and learned reconstruction. Performance obtained for one conductivity model, electrode layout, implantation pattern, pathology, state, or task does not establish equivalent performance outside that regime. The training and validation distributions form part of the evidential boundary of any reconstruction claim. ESI and virtual iEEG consequently occupy distinct but complementary roles: ESI provides an explicit anatomical coordinate system, forward-model discipline, and established localisation baselines, whereas virtual iEEG targets iEEG-indexed events, features, representations, or waveforms. Source-imaging baselines can test whether learned inference adds information beyond anatomy-aware non-invasive estimation, while successful localisation alone remains insufficient evidence for contact-level waveform reconstruction.

\section{Hidden-event inference and event-centred representations}

Studies in this area span the event-inference and feature-translation categories defined in Fig.~\ref{fig:framework}(b), with each target imposing a different evidential burden. The variables \(\boldsymbol{x}\) and \(c\) denote a scalp EEG segment and an iEEG-defined event or state, respectively. Hidden-event inference estimates \(\boldsymbol{p}(c\mid \boldsymbol{x})\), for which success depends on a reproducible scalp correlate of that state. Feature translation instead examines whether scalp data preserve an iEEG-derived representation. Event discrimination and representation-level agreement require target-specific evidence. Neither endpoint quantifies continuous contact-level waveform recovery.

\subsection{Scalp correlates of intracranial seizures}

Early seizure studies examined whether selected intracranial events leave detectable distributed scalp correlates. Lam et al. analysed 25 scalp-negative mesial temporal seizures from 10 patients and approximately 189 hours of control recordings from 13 patients~\cite{lam2016widespread}. Their leave-one-patient-out coherence detector identified eight prespecified seizures and one previously overlooked event. It detected seizures in 40\% of affected patients and lateralised every detected event correctly. The corrected false-alarm rate was 0.31 per day, with 75\% positive predictive value. The detected events were dominated by sleep, and three of four initial false alarms occurred around sleep--wake transitions. Principal components were derived from combined seizure and control records before cross-validation, while final features were selected using aggregate cross-validation performance. The findings provide evidence for widespread, state-dependent scalp signatures of some focal seizures; focal foramen-ovale waveform reconstruction was not evaluated.

SCOPE-mTL examined the related but narrower task of identifying occult onset before an already visible scalp seizure~\cite{Lam2017}. In 24 patients, foramen-ovale onset preceded the scalp ictal pattern by \(14.3\pm10.4\) seconds on average. Among 68 seizures with a lead exceeding five seconds, 50 were identified as early, yielding 73\% sensitivity. Predicted onset time also correlated with the invasive reference (\(r=0.69\)). Reviewers lateralised only 34 of all 86 seizures, although 32 of those 34 decisions were correct. The often-cited 94\% value is the conditional accuracy among lateralised cases rather than the overall lateralisation rate. Because the analysis began from a known scalp ictal onset and required expert assessment, the evaluation concerned assisted retrospective annotation rather than prospective detection.

Deep-learning models have extended occult-onset classification to larger event sets. Li et al. analysed 102 mesial-temporal seizures from 19 patients using graph-based CNN and LSTM models~\cite{Li2021}. Five repeated patient-disjoint splits each included 16 training patients and three test patients. Accuracy exceeded 97\% on sampled one-second epochs, including scalp-negative seizure portions, and lateralisation results were also reported. Epoch-level accuracy does not provide an event-level false-alarm rate or forecasting performance during continuous monitoring, because even a sub-1\% non-seizure epoch error can accumulate over hours. The ten-minute pre-seizure label may also capture peri-ictal state or temporal-context cues. The study therefore extends evidence for occult-onset classification, while event-wise prospective performance remains unresolved.

\subsection{Intracranial IED inference across evaluation settings}

The contrast between balanced segments and continuous monitoring is also evident in interictal epileptiform discharge (IED) detection. Spyrou et al. studied 18 patients with approximately 20 minutes of concurrent scalp and foramen-ovale EEG~\cite{Spyrou2016IED}. On balanced 325-ms segments centred on invasive IED peaks, cross-patient accuracy was 65\% and the area under the receiver-operating-characteristic curve (AUC) was 0.67. During continuous application, a threshold producing 3.6 false detections per minute recovered only a minority of high-certainty scalp-invisible events. The balanced-segment result thus exceeded the detection yield obtained under the class distribution of continuous recordings.

HEAnet increased both cohort size and recording duration, analysing 8,395 hours of simultaneous data from 51 patients and 22,762 expert-labelled positive events~\cite{AbouJaoude2022}. Event-level AUC reached 0.89, whereas precision--recall AUC was 0.39, reflecting the operational effect of class imbalance. At positive predictive value (PPV) near 0.7, sensitivity was \(0.25\pm0.08\) and specificity was \(0.996\pm0.002\). The false-detection rate was \(0.86\pm0.34\) per minute. Two independent scalp-only cohorts yielded patient-level diagnostic AUCs of 0.88 and 0.95, with lateralisation accuracies of 100\% and 92\%. These cohorts lacked simultaneous iEEG, so their endpoints validate diagnostic and lateralisation proxies rather than event-level hippocampal activity. Sensitivity at the same high-precision threshold also fell to 0.06 during wakefulness, indicating substantial state dependence.

Phase-based analyses identify propagation as another determinant of event observability~\cite{Pyrzowski2021}. In 16 patients with an average of 64.8 hours of simultaneous data, 79\% of intracranial channels showed a significant scalp zero-crossing pattern. The proportion was 91\% for mesial-temporal channels but 16\% for events classified as non-propagating. A bank of 101 selected templates distinguished 19 patients with temporal-lobe epilepsy from 27 controls, with a median AUC of 0.72 compared with approximately 0.59 for visual assessment. Automated invasive labels had an estimated precision of only 72\%, cross-patient templates involved 9 of 16 source patients, and the external endpoint was diagnosis rather than invasive event confirmation. The 91\%-to-16\% contrast provides evidence that propagation contributes to scalp observability, although label imprecision and template selection constrain the generalisability of this association.

Other studies have targeted state- or spectrum-level markers rather than individual events. De Stefano et al. found that scalp-invisible hippocampal IEDs coincided with increased scalp delta power and lateralising slow-power topographies in nine patients~\cite{DeStefano2022}. The delta increase occurred at frequencies of at least 1.4 Hz. Each analysis selected one frequent IED type and recording state per patient and averaged power spectra. The study consequently demonstrated an event-locked population perturbation, while single-event detection was outside its evaluation. Yamaguchi et al. subsequently used 58 scalp features in 17 patients~\cite{Yamaguchi2026}. Repeated-holdout accuracy was 0.70 and AUC was 0.78 for occult mesial-temporal IED epochs. The 8:2 epoch split nevertheless placed data from the same patients in training and test sets. External cohorts of 35 patients with mesial-temporal epilepsy and 33 non-epileptic controls validated proxies based on the frequency of classified epochs without simultaneous iEEG. Across these studies, the evidence progresses from averaged biomarkers to within-cohort epoch classification; externally verified event detection has not yet been demonstrated.

\subsection{Event-centred intracranial representations}

Event-centred representation methods change the estimand from a label to an iEEG-informed feature or signal representation. A historical precursor was provided by Kaur et al., who studied six patients performing a P300 matrix-speller in non-concurrent pre-implant scalp and post-implant iEEG sessions~\cite{kaur2014empirical}. The transformation ran from iEEG event-related potentials (ERPs) to scalp ERPs rather than from scalp EEG to iEEG. Four patients achieved broadly comparable BCI performance across modalities when their ERPs remained stable. Session instability reduced transformed performance to 19\% and 25\% in the other two patients. The experiment provides evidence for a task-locked cross-modal relationship and shows its sensitivity to changes in state, session, and coverage.

Early paired representation studies used highly constrained event-centred windows. The coupled-dictionary study analysed iEEG-marked 325-ms IED windows from 10 patients with patient-specific random train--test splits~\cite{Spyrou2016CDLSA}. Normalised test errors ranged from 0.067 to 0.187, but the number of atoms was partly chosen after inspection of test error. Antoniades et al. learned a scalp-to-intracranial representation mapping in the 18-patient King's College cohort~\cite{Antoniades2018}. Their asymmetric--symmetric autoencoder improved leave-patient-out IED classification from 62\% for linear regression and 65\% for direct time--frequency features to 68\%. Waveform validation was limited to rejecting the null hypothesis of zero correlation; absolute correlation, error, and spectral fidelity were not reported. Mapper selection also used scalp data from the held-out patient. Under these protocols, iEEG-informed representations retained discriminative information in event-centred settings. Continuous contact-level waveform reconstruction was not evaluated.

Coupled dictionary learning formalises the shared-code assumption by representing scalp and intracranial observations with common sparse coefficients,
\begin{equation}
\begin{aligned}
\boldsymbol{x}&\approx \boldsymbol{D}_s\boldsymbol{A},\qquad \boldsymbol{y}\approx \boldsymbol{D}_i\boldsymbol{A},\\
\boldsymbol{A}^\ast&=\arg\min_{\boldsymbol{A}}\|\boldsymbol{x}-\boldsymbol{D}_s\boldsymbol{A}\|_2^2+\lambda\|\boldsymbol{A}\|_1,
\end{aligned}
\label{eq:coupled_dictionary}
\end{equation}
so that \(\hat{\boldsymbol{y}}=\boldsymbol{D}_i\boldsymbol{A}^\ast\). Autoencoders replace the explicit sparse code with \(\hat{\boldsymbol{y}}=\boldsymbol{D}_\theta(E_\phi(\boldsymbol{x}))\). These formulations expose the assumption that an aligned scalp code supports the corresponding intracranial representation. Empirical evaluation of that assumption has remained restricted to the event windows and anatomies represented during training.

Tensor and common-feature methods address related but distinct targets. The Tucker/canonical polyadic decomposition (CPD) model projected scalp time--frequency data through factors derived from invasive IEDs~\cite{Abdi2021CPD}. Accuracy was 84.2\% within patients and 72.6--72.7\% between patients, with near-zero sensitivity for some held-out patients. SCA-IEDP instead weighted scalp decompositions using one expert's iEEG-based label certainty~\cite{Abdi2021SCA}, whereas SCFA learned features shared across scalp IED segments~\cite{Abdi2022SCFA}. Their between-patient accuracies were approximately 63--68\%, often with marked sensitivity--specificity imbalance. The Tucker/CPD method therefore provides an iEEG-informed feature projection, while SCA-IEDP and SCFA provide target-informed scalp detection. None of these evaluations included contact-level waveform reconstruction.

Independent replication is further limited by cohort reuse. Spyrou's detector, the Antoniades autoencoders, and the SCA, CPD, and SCFA studies all analyse variants of one 18-patient foramen-ovale cohort. Their architectural comparisons provide within-cohort evidence, but remain repeated analyses of a single clinical sample.

Early event-inference and representation studies collectively support inference within a relatively narrow domain. Most used private foramen-ovale or temporal-lobe cohorts with limited contact coverage, few patients, and downstream IED-detection endpoints. These conditions define the anatomical, clinical, and evaluative scope of the reported results. Evidence is strongest for detecting selected intracranial events and learning event-centred intracranial representations within the sampled settings; recovery across arbitrary continuous contact-level waveforms, patients, regions, and tasks remains untested. Explicit waveform reconstruction estimates time-resolved intracranial signals, for which waveform fidelity, predictive uncertainty, and generalisation require separate evaluation.

\section{Waveform reconstruction, uncertainty, and generalisation}

Waveform reconstruction extends the target hierarchy in Fig.~\ref{fig:framework}(b) from event- or feature-level inference to contact-level signal estimation, with corresponding requirements for fidelity, uncertainty, and utility in Fig.~\ref{fig:framework}(c). Its evidential scope also depends on the information available before inference. Figure~\ref{fig:generalisation_regimes} distinguishes patient-specific evaluation, adapted cross-patient evaluation using limited target-patient iEEG, and zero-shot evaluation without target-patient invasive recordings. Variational autoencoders (VAEs), GANs, normalising flows, and diffusion models provide different parameterisations of deterministic or probabilistic reconstruction~\cite{Kingma2013,Goodfellow2014,Isola2017,Dinh2016,Durkan2019,Ho2020,Song2021,zhang2026ddlsr}, but architecture alone does not determine evidential strength. Pairing, conditioning, target selection, uncertainty calibration, the held-out unit, and access to invasive information remain decisive. Non-stationarity, reference dependence, heterogeneous electrode layouts, and the possibility that multiple local signals are compatible with the same scalp observation further constrain contact-level identifiability.

\subsection{Cross-modal correspondence and conditional waveform reconstruction}

Simultaneous and population-level comparisons establish that scalp and intracranial recordings share selected task- and frequency-dependent structure without implying contact-waveform equivalence. Boran et al. reported workload-related hippocampal--scalp theta/alpha phase locking in eight of nine patients during verbal working memory~\cite{Boran2019}. In a 12-patient study covering 965 bipolar SEEG sites, Barborica et al. found stronger and more sustained memory decoding from SEEG, with neither beamforming nor independent-component analysis improving scalp decoding~\cite{Barborica2023}. Janiukstyte et al. compared source-localised scalp maps from 17 controls with iEEG maps from 234 different patients and observed frequency-dependent divergence, including negative scalp--iEEG similarity in gamma (\(\rho=-0.42\))~\cite{Janiukstyte2023}. Because that comparison was unpaired, it characterises a population-level modality difference rather than patient-specific translation. Collectively, these studies support partial cross-modal correspondence while indicating substantial modality-specific information.

Explicit contact-level reconstruction can be formalised using paired, contact-indexed examples \(\{(\boldsymbol{x}_n,\boldsymbol{y}_{n,i},\boldsymbol{A}_n,\boldsymbol{r}_{n,i},\boldsymbol{\mit\Omega}_n)\}_{n=1}^{N},\) where \(\boldsymbol{x}_n\in\mathbb{R}^{C_s\times T}\) denotes a scalp EEG segment, \(\boldsymbol{y}_{n,i}\in\mathbb{R}^{T}\) denotes the simultaneous iEEG waveform at target contact \(i\), \(\boldsymbol{A}_n\) denotes anatomical information, \(\boldsymbol{r}_{n,i}\) denotes the geometry and anatomical location of the target contact, and \(\boldsymbol{\mit\Omega}_n\) denotes acquisition and deployment context, including montage, hardware, state, task, and domain information. A deterministic contact-conditioned model estimates
\begin{equation}
\hat{\boldsymbol{y}}_{n,i}=f_\theta(\boldsymbol{x}_n,\boldsymbol{A}_n,\boldsymbol{r}_{n,i},\boldsymbol{\mit\Omega}_n),
\label{eq:deterministic_mapping}
\end{equation}
where \(\boldsymbol{x}_n\) provides concurrent scalp information, \(\boldsymbol{A}_n\) and \(\boldsymbol{r}_{n,i}\) provide anatomical and contact-specific conditioning, \(\boldsymbol{\mit\Omega}_n\) represents contextual information, and \(\theta\) captures cohort-level regularities.

For explicit waveform estimates, the training objective determines which aspects of the target are prioritised. A waveform-oriented objective may combine time-domain, spectral, and correlation terms:
\begin{equation}
\begin{aligned}
\mathcal{L}_{\mathrm{rec}}={}&\lambda_t\|\boldsymbol{y}-\hat{\boldsymbol{y}}\|_1\\
&+\lambda_f\|\log(\epsilon_0+|\mathcal{S}\boldsymbol{y}|)-\log(\epsilon_0+|\mathcal{S}\hat{\boldsymbol{y}}|)\|_1\\
&+\lambda_\rho(1-\rho(\boldsymbol{y},\hat{\boldsymbol{y}})),
\end{aligned}
\label{eq:rec_loss}
\end{equation}
where \(\mathcal{S}\) denotes a short-time Fourier transform, \(\epsilon_0>0\) ensures numerical stability, and \(\rho\) is a contact-wise correlation. Event-oriented, feature-translation, and decoder-oriented models instead optimise event, latent-space, or downstream-task objectives. Performance under one objective therefore does not quantify fidelity for another estimand.

E2SGAN made waveform-domain translation explicit using concurrent pre-ictal EEG and SEEG from seven patients~\cite{Hu2022E2SGAN}. Signals were downsampled to 64 Hz and segmented into highly overlapping 15.875-s windows, with scalp--depth channel pairs selected using physical distance and spectral Hellinger similarity. A conditional Wasserstein GAN combining magnitude and instantaneous-frequency representations improved transformed dynamic-time-warping and spectral-distance measures relative to raw EEG and several generative baselines. Because pairing preferentially retained scalp--SEEG channels with pre-existing similarity, the demonstrated synthesis concerned an observable subset rather than arbitrary contacts. Absolute contact-level correlations, event-level endpoints, and clinical utility were not reported, and the sampling rate restricted evaluation to frequencies below 32 Hz.

Adversarial and variational models were subsequently evaluated on event-centred segments from the established 18-patient foramen-ovale cohort. A deterministic U-Net generator trained on 320-ms iEEG-centred segments achieved 68\% leave-patient-out IED classification accuracy without a quantitative waveform-fidelity metric~\cite{Abdi2023GAN}. A variational-autoencoder--conditional-GAN extension reported within-patient mean-squared error of 0.014, Pearson correlation of 0.35, and cosine similarity of 0.34, with between-patient classification accuracy of 69\%~\cite{Abdi2025VAE}. Repeated-sample diversity, cross-patient waveform fidelity, and predictive-interval coverage were not evaluated. The results therefore support event-centred generation of iEEG-like waveforms within this cohort, while dependence on the concurrent scalp segment and predictive uncertainty remain incompletely characterised.

These adversarial and variational studies belong to the same 18-patient cohort lineage used by earlier detection and representation models. Their architectural comparisons therefore constitute repeated analyses of a single clinical sample rather than independent replications. The recurring decline from within-patient to between-patient performance is consistent with sensitivity to patient-specific morphology or implantation geometry, although the available experiments do not isolate either factor.

CAST extended contact-conditioned reconstruction to pooled cross-patient training under the adapted regime in Fig.~\ref{fig:generalisation_regimes}(b)~\cite{Pham2026CAST}. Nine working-memory patients and 14 retained visual-task patients contributed 1,282 contacts, with the temporal encoder trained under leave-one-patient-out cross-validation. Each test patient additionally contributed 20\% of simultaneous scalp--iEEG data, corresponding to approximately 2--12 min, for decoder adaptation and observable-contact selection. Among contacts retained using \(r\geq0.15\) on the adaptation set, mean correlations across all patients were 0.345 and 0.323 in the two cohorts. Applying the stated criterion of at least ten selected contacts to the reported subject-level results yields unweighted means of 0.384 for 7/9 patients and 0.416 for 10/14 patients, whereas the reported aggregate values of 0.458 and 0.545 do not reproduce under that criterion. The reported precentral maximum of \(r=0.864\) was based on two contacts. These results demonstrate adapted cross-patient reconstruction with a performance--coverage trade-off, but do not establish zero-shot pre-implant generalisation.

The conditional adversarial objective makes explicit both the information supplied to the generator and the criterion applied to generated signals. For notational simplicity, anatomical, contact-geometric, and contextual conditioning variables are suppressed in the generic objectives below. For a generator \(\mathcal{G}_\theta\) and discriminator \(\mathcal{D}\),
\begin{equation}
\begin{aligned}
\min_{\mathcal{G}}\max_{\mathcal{D}}\;&\mathbb{E}_{\boldsymbol{x},\boldsymbol{y}}[\log \mathcal{D}(\boldsymbol{x},\boldsymbol{y})]\\
&+\mathbb{E}_{\boldsymbol{x},\boldsymbol{z}}[\log(1-\mathcal{D}(\boldsymbol{x},\mathcal{G}(\boldsymbol{x},\boldsymbol{z})))]\\
&+\lambda\|\boldsymbol{y}-\mathcal{G}(\boldsymbol{x},\boldsymbol{z})\|_1 .
\end{aligned}
\label{eq:conditional_gan}
\end{equation}
The reconstruction and adversarial terms balance paired fidelity against realistic intracranial morphology. Weak adversarial weighting may favour smoothed estimates, whereas excessive weighting can reward plausible iEEG-like outputs that are only weakly constrained by the concurrent scalp segment. Conditional VAEs instead maximise an evidence lower bound,
\begin{equation}
\begin{aligned}
\mathcal{L}_{\mathrm{VAE}}=&\;
\mathbb{E}_{q_\phi(\boldsymbol{z}\mid \boldsymbol{x},\boldsymbol{y})}[\log p_\theta(\boldsymbol{y}\mid \boldsymbol{x},\boldsymbol{z})]\\
&-\mathrm{KL}\!\left(q_\phi(\boldsymbol{z}\mid \boldsymbol{x},\boldsymbol{y})\|p_\theta(\boldsymbol{z}\mid \boldsymbol{x})\right),
\end{aligned}
\label{eq:cvae}
\end{equation}
allowing one-to-many structure to be represented through the latent variable \(\boldsymbol{z}\). Neither stochastic latent variables nor adversarial realism alone establish that predictions depend materially on the concurrent scalp input or that the resulting uncertainty is calibrated.

Across explicit waveform-reconstruction studies, pairing rules, event selection, target-patient adaptation, contact retention, and cohort-specific priors differ substantially. These choices influence both reported fidelity and the supported generalisation claim. Adversarial and variational objectives distinguish paired waveform similarity from realistic intracranial morphology, but neither objective by itself establishes scalp-conditioned recovery; corresponding conditioning controls and held-out evaluation remain necessary.

\subsection{Probabilistic reconstruction and predictive uncertainty}

Probabilistic reconstruction estimates a conditional distribution \(p_\theta(\boldsymbol{y}_i\mid \boldsymbol{x},\boldsymbol{A},\boldsymbol{r}_i,\boldsymbol{\mit\Omega})\) rather than a single waveform, allowing residual ambiguity in the scalp-to-intracranial mapping to be represented explicitly~\cite{Dinh2016,Kingma2018,Durkan2019}. In a conditional normalising flow, an invertible transform \(\boldsymbol{u}=f_\theta(\boldsymbol{y}_i;\boldsymbol{x},\boldsymbol{A},\boldsymbol{r}_i,\boldsymbol{\mit\Omega})\) maps the target iEEG to a simple base density:
\begin{equation}
\log p_\theta(\boldsymbol{y}_i\mid \boldsymbol{x},\boldsymbol{A},\boldsymbol{r}_i,\boldsymbol{\mit\Omega})=\log \boldsymbol{p}(\boldsymbol{u})
+\log\left|\det\frac{\partial f_\theta(\boldsymbol{y}_i;\boldsymbol{x},\boldsymbol{A},\boldsymbol{r}_i,\boldsymbol{\mit\Omega})}{\partial \boldsymbol{y}_i}\right|.
\label{eq:conditional_flow}
\end{equation}
The likelihood provides a distributional objective, but its evidential value depends on whether the learned conditional distribution is constrained by the concurrent scalp segment and whether its predictive uncertainty is calibrated.

\begin{figure*}[!t]
    \centering
    \includegraphics[width=\textwidth]{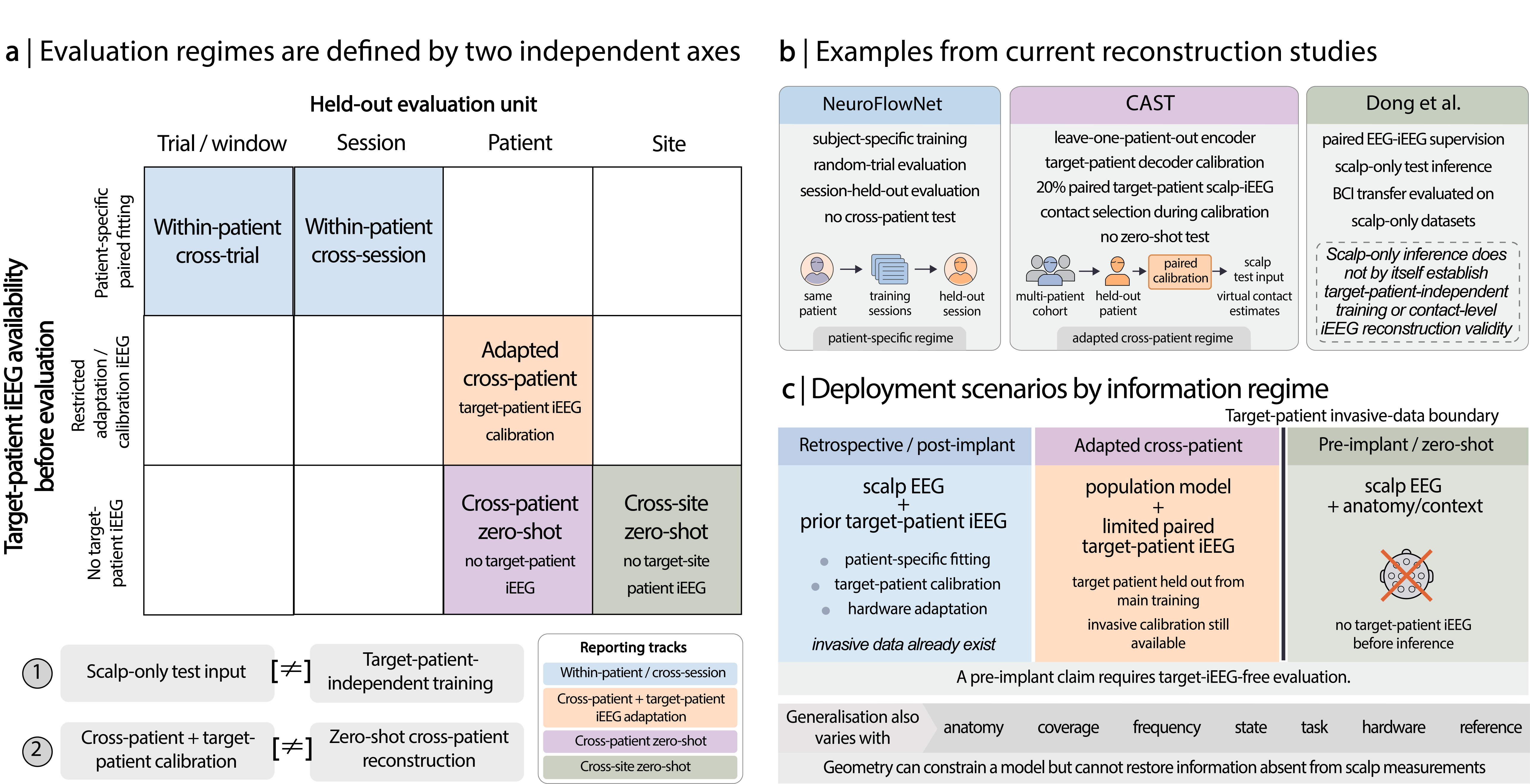}
    \caption{\textbf{Generalisation regimes defined by held-out units and target-patient iEEG availability.}
    \textbf{(a)} Held-out unit and prior target-patient iEEG jointly distinguish within-patient, adapted cross-patient, and target-iEEG-free zero-shot evaluation; scalp-only test input does not itself establish zero-shot generalisation.
    \textbf{(b)} NeuroFlowNet~\cite{He2026NeuroFlowNet}, CAST~\cite{Pham2026CAST}, and Dong et al.~\cite{Dong2026Bridge} illustrate patient-specific, adapted cross-patient, and scalp-only inference after paired supervision, respectively.
    \textbf{(c)} Deployment ranges from retrospective settings permitting target-patient iEEG to pre-implant zero-shot use without invasive adaptation; anatomical conditioning cannot restore information absent from scalp measurements.}
    \label{fig:generalisation_regimes}
\end{figure*}

NeuroFlowNet applied conditional normalising flows to band-limited temporal-lobe reconstruction under the patient-specific regime in Fig.~\ref{fig:generalisation_regimes}(a)~\cite{He2026NeuroFlowNet}. Separate models were trained for three selected patients from the nine-patient public working-memory cohort. Mean temporal correlation was approximately 0.50 under random cross-trial testing and 0.54 with held-out sessions, while regional cross-session correlations ranged from approximately 0.05 to 0.84. Cross-patient reconstruction was not evaluated. More importantly for the probabilistic claim, nominal 90\% predictive intervals covered only approximately 38--43\% of observations. Predictive spread was associated with some reconstruction errors, but the severe undercoverage indicates that uncertainty magnitude and uncertainty calibration are distinct properties. The reported evidence therefore supports patient-specific, band-limited probabilistic reconstruction with substantial regional variation, rather than calibrated uncertainty across patients or broader anatomy.

Diffusion models provide an alternative representation of complex conditional target distributions through iterative denoising. A typical contact-conditioned objective is
\begin{equation}
\mathcal{L}_{\mathrm{diff}}=
\mathbb{E}_{\boldsymbol{y}_{i,t},t,\boldsymbol{x},\boldsymbol{A},\boldsymbol{r}_i,\boldsymbol{\mit\Omega},\boldsymbol{\epsilon}}
\|\boldsymbol{\epsilon}-\boldsymbol{\epsilon}_\theta(\boldsymbol{y}_{i,t},t,\boldsymbol{x},\boldsymbol{A},\boldsymbol{r}_i,\boldsymbol{\mit\Omega})\|_2^2,
\label{eq:diffusion}
\end{equation}
where \(\boldsymbol{y}_{i,t}\) is a noised intracranial signal at target contact \(i\), \(\boldsymbol{\epsilon}\sim\mathcal{N}(\boldsymbol{0},\boldsymbol{I})\), and \(t\) indexes the diffusion step. The conditioning variables include scalp evidence together with anatomical, contact-geometric, and deployment information. As with adversarial and variational objectives, successful denoising does not by itself establish how strongly the generated waveform depends on the concurrent scalp segment rather than the learned intracranial prior.

Dong et al. combined pretrained EEG representations, geometry-derived constraints, and conditional diffusion using paired scalp EEG--iEEG supervision, followed by scalp-only inference~\cite{Dong2026Bridge}. On the paired reconstruction benchmark, mean waveform correlation was approximately 0.38, power-spectral similarity approximately 0.51, and phase consistency approximately 0.52. Across 12 EEG-only BCI datasets, the generated representations produced a mean F1 improvement of 5.2\%. The primary reconstruction comparator was noise-guided generation, and predictive-interval calibration was not evaluated. Because the BCI datasets lacked simultaneous iEEG targets, the downstream gains provide evidence of representation or decoder utility rather than interval-specific intracranial fidelity. The waveform and BCI endpoints therefore support different claims and should not be interpreted as interchangeable evidence of probabilistic reconstruction.

Uncertainty affecting virtual iEEG arises at different levels. \textit{Label uncertainty} concerns ambiguity or disagreement in the invasive reference; SCA-IEDP incorporated this source by weighting IEDs according to one expert's label certainty~\cite{Abdi2021SCA}. \textit{Aleatoric uncertainty} concerns residual variability in \(\boldsymbol{Y}\) conditional on the available scalp, anatomical, and contextual information, including cases in which multiple intracranial signals remain compatible with the same observation. \textit{Epistemic uncertainty} reflects limited support for model parameters under unfamiliar patients, anatomies, artefacts, sites, or states. These sources should not be inferred solely from the presence of latent variables, generator noise, ensemble dispersion, or an explicit likelihood.

Evaluation of predictive uncertainty requires both calibration and discrimination. Calibration tests whether nominal predictive intervals attain their stated empirical coverage, whereas discrimination tests whether larger estimated uncertainty is associated with larger realised error. Proper scoring rules, including negative log likelihood and the continuous ranked probability score, assess the full predictive distribution and complement interval coverage and width~\cite{Gneiting2007}. Results can additionally be stratified by patient, contact, anatomy, frequency, and event type to identify systematic regions of overconfidence or poor predictability. Risk--coverage or selective-prediction curves further test whether abstaining on uncertain targets reduces reconstruction error.

A probabilistic output also requires evidence of conditional dependence on the concurrent scalp segment. Realistic or well-scored samples can arise partly from anatomical, patient, task, or marginal iEEG priors without accurately tracking the corresponding intracranial interval. Time-shifted or shuffled scalp--iEEG pairs, scalp perturbation or ablation, and anatomy-mismatched targets provide direct tests of this dependence. Such controls distinguish distributional realism from scalp-conditioned accuracy, while the broader requirements for leakage control, split independence, and deployment transfer are addressed in the subsequent generalisation and validation sections.

\subsection{Generalisation and target-patient adaptation}

Generalisation is defined jointly by the independent unit held out during evaluation and the amount of target-patient iEEG available before that evaluation, as summarised in Fig.~\ref{fig:generalisation_regimes}. Cross-trial, cross-session, cross-patient, and cross-site splits test progressively different sources of distributional change but do not form a single hierarchy of difficulty. At the patient level, patient-specific fitting, cross-patient evaluation with restricted target-patient iEEG adaptation, and target-iEEG-free zero-shot evaluation support distinct claims. Scalp-only inference specifies the information supplied at test time; it does not establish that the target patient was absent from paired fitting, calibration, or target selection.

Current reconstruction studies occupy different positions within this framework. NeuroFlowNet used patient-specific training and evaluated transfer across trials and sessions within the same patient~\cite{He2026NeuroFlowNet}. CAST excluded each target patient from the main encoder training but used paired target-patient scalp--iEEG data for decoder adaptation and contact selection~\cite{Pham2026CAST}; its results therefore support adapted cross-patient reconstruction rather than pre-implant zero-shot inference. Dong et al. used paired EEG--iEEG supervision followed by scalp-only test inference~\cite{Dong2026Bridge}. These protocols differ in prior exposure to invasive information even when scalp EEG is the only signal supplied during final inference, and their reported performances consequently represent different generalisation claims.

Anatomical transfer introduces an additional constraint. Contacts differ in location, depth, tissue proximity, orientation, reference neighbourhood, and implantation geometry, all of which affect both the measured intracranial signal and its scalp observability. A patient-level split therefore does not by itself establish generalisation to unseen regions or implantation configurations. Anatomical descriptors can condition reconstruction and support stratified evaluation, but they cannot restore components absent from the scalp measurement. Reporting performance across regions, depths, frequencies, and all eligible versus selected contacts is consequently necessary to distinguish broad transfer from favourable anatomical coverage.

Cross-study evidence also depends on independence of the underlying data. Repeated analyses of overlapping patients, contacts, or sessions can support methodological comparison without constituting independent replication. Generalisation benchmarks should therefore align the split with the intended deployment setting, isolate any target-patient adaptation data from evaluation, and retain a separate target-iEEG-free track when zero-shot claims are made. Patient-level aggregation and explicit reporting of target coverage preserve the relevant independent unit, whereas pooled-window performance can obscure between-patient variation. Table~\ref{tab:study_evidence} summarises representative studies by evaluation design, principal result, and the boundary of the supported claim; direct ranking is not appropriate because targets, cohorts, adaptation regimes, and denominators differ.

\begin{table*}[!t]
\caption{Representative evidence for scalp-to-intracranial inference, including cohort design, principal results, and boundaries of the supported claims}
\centering
\footnotesize
\begin{tabularx}{\linewidth}{p{0.12\linewidth}p{0.20\linewidth}p{0.25\linewidth}X}
\toprule
\textbf{Study / Model} & \textbf{Cohort and estimand} & \textbf{Principal result} & \textbf{Evidence boundary} \\
\midrule
Lam et al.~\cite{lam2016widespread} & 10 patients contributing 25 scalp-negative mesial-temporal seizures plus 13 patients contributing control records; leave-one-patient-out network detection & Eight of the 25 prespecified seizures plus one previously overlooked seizure detected; three false alarms; 0.31 false alarms/day; PPV 75\% & PCA fitted to all seizure and control records before cross-validation; feature set selected on cross-validation performance; 8/9 detections occurred during sleep; local waveform recovery not evaluated \\
\midrule
Abou Jaoude et al.~\cite{AbouJaoude2022} & \(n=51\); 8,395 h simultaneous scalp--foramen-ovale EEG; patient-stratified fivefold cross-validation; hidden hippocampal-activity detection & ROC AUC \(=0.89\); precision--recall AUC \(=0.39\); at PPV \(=0.70\): sensitivity \(=0.25\), false-positive rate \(=0.86\)/min & Primary training and expert-labelled testing used sleep; external diagnosis/lateralisation cohorts were scalp-only; awake sensitivity \(=0.06\) at the sleep-derived PPV \(\simeq0.9\) threshold using FOnet labels \\
\midrule
Antoniades et al.~\cite{Antoniades2018} & Simultaneous scalp--foramen-ovale cohort (\(n=18\)); 325-ms IED/non-IED trials; leave-subject-out pseudo-iEEG classification & IED accuracy: 62\% linear mapping; 65\% direct time--frequency scalp features; 68\% asymmetric-symmetric-autoencoder representation & The test subject's scalp data selected the most-correlated of 17 subject-specific mappers; waveform comparison was qualitative; absolute correlation and waveform error were not reported \\
\midrule
E2SGAN~\cite{Hu2022E2SGAN} & \(n=7\); simultaneous pre-ictal EEG--SEEG; 15.875-s windows with 75\% overlap; leave-one-patient-out evaluation; preselected channel pairs & \(\log_2\) distance ratios versus raw EEG: DTW \(-0.414\), PSD \(-1.480\), Hellinger \(-0.221\); best PSD and Hellinger results among compared models & Pair selection used distance intervals and signal similarity; contact-wise correlation and event endpoints were not reported; downsampling to 64 Hz imposed a 32-Hz Nyquist limit; separation relative to overlapping-window generation was not reported \\
\midrule
VAE-cGAN~\cite{Abdi2025VAE} & Reused simultaneous cohort (\(n=18\)); 320-ms IED/non-IED segments; intra-subject 70:10:20 split and inter-subject ensemble & Intra-subject mean-squared error 0.014, Pearson \(r=0.35\), cosine 0.34; inter-subject IED accuracy 69\% versus 68\% for ASAE and cGAN & Waveform metrics were intra-subject only; inter-subject models were restricted to training subjects with intra-subject accuracy \(>70\%\); cross-subject waveform fidelity, diversity, interval coverage, and uncertainty calibration were not evaluated \\
\midrule
NeuroFlowNet~\cite{He2026NeuroFlowNet} & Three selected patients from a nine-patient simultaneous dataset; subject-specific medial-temporal reconstruction; random-trial 90:10 and session-held-out tests & Mean \(r\): random-trial 0.50; session-held-out 0.54; regional session-held-out 0.05--0.84 & No cross-patient test; spectral evaluation was mainly 0.5--50 Hz and excluded high-gamma/HFO claims; validation-set coverage of nominal 90\% intervals was 38.1--43.2\% \\
\midrule
CAST~\cite{Pham2026CAST} & Cohorts \(n=9,14\); 502/780 contacts; leave-one-subject-out encoder plus target-patient decoder calibration & All-contact mean \(r\): 0.271/0.274; observable-contact mean \(r\) across all patients: 0.345/0.323; applying the stated \(N_{\mathrm{obs}}\geq10\) criterion to subject-level Tables 5--6 gives viable-subject mean \(r\): 0.384/0.416 & Target-patient calibration used 20\% paired scalp--iEEG data and selected contacts at calibration \(r\geq0.15\); evaluation used the remaining 80\%; the paper's reported viable-subject aggregates were not used because they were inconsistent with the stated eligibility criterion; no zero-shot test \\
\midrule
Dong et al.~\cite{Dong2026Bridge} & Paired conditional-diffusion benchmark with scalp-only test inference; 12 scalp-only BCI datasets & Paired benchmark: Pearson correlation 0.38, PSD similarity 0.51, phase consistency 0.52; downstream F1 gain 2.8--8.6\% (mean 5.2\%) & Primary reconstruction comparison was against noise-guided generation; predictive uncertainty calibration was not evaluated; BCI datasets lacked simultaneous iEEG targets, so downstream gain does not establish interval-specific reconstruction \\
\bottomrule
\end{tabularx}
\label{tab:study_evidence}
\end{table*}

\subsection{Cross-modal transfer and reconstruction boundaries}

Cross-modal pretraining and transfer address related but distinct objectives from concurrent STIR. Mixed-modality pretraining learns reusable representations from marginal EEG and iEEG corpora, whereas scalp-pretrained ECoG transfer uses non-invasive data to initialise a model whose deployment input remains invasive. EpiNT pretrained channel-independent masked-autoencoder and vector-quantised representations on 13 datasets comprising 1,199 patients and more than 2,700 recording hours~\cite{zhang2025cross}. Mixed EEG--iEEG pretraining produced the strongest overall performance across six retrospective classification tasks, with the highest F1 among compared pretrained models on four tasks under full-parameter tuning. Performance varied with tuning depth, and training from scratch was superior on one CHB-MIT seizure-prediction task. Because the pretraining corpus was unpaired and modality ablations were not matched for data volume, these results support reusable cross-modal representations rather than patient-specific scalp--intracranial correspondence or a modality-invariant physiological latent space.

CORTEG examined a different transfer direction: scalp-EEG pretraining supplied the backbone and spatial codebook, but ECoG remained the input at inference~\cite{Yang2026CORTEG}. Finger-trajectory and audio-envelope decoding were evaluated in nine and 16 patients, respectively. Held-out-patient performance relied on target-patient ECoG fine-tuning, whereas zero-shot correlations were approximately 0.09 and 0.11. The reported gains therefore concern sample-efficient adaptation of invasive decoders rather than generation of intracranial activity from concurrent scalp EEG.

These studies broaden the evidence for representations that transfer across recording modalities, but their information flow differs fundamentally from virtual iEEG waveform reconstruction. Unpaired pretraining can demonstrate reusable statistical structure without establishing paired correspondence, while invasive-input fine-tuning can improve decoder performance without requiring scalp-to-intracranial prediction at deployment. Representation transfer, waveform fidelity, and downstream utility consequently constitute separate claims and require evaluation against their respective targets and information regimes.

\section{Validation and translational evidence}
Validation of virtual iEEG requires three distinct questions: whether the output agrees with the invasive target, whether that agreement depends on the concurrent scalp input, and whether the relationship persists under the intended deployment conditions. \textit{Measurement validity} concerns agreement with the simultaneous, operationally defined invasive target under the stated estimand. \textit{Conditional validity} concerns the information contributed by concurrent scalp EEG beyond anatomy, task context, target-patient adaptation data, and the marginal structure of iEEG. \textit{Deployment validity} concerns persistence across the intended patients, sites, states, hardware, and adaptation regimes. Reconstruction claims require measurement and conditional validity, whereas translational claims additionally require deployment validity and incremental utility. Downstream performance without measurement or conditional validation may demonstrate a useful representation but does not establish contact-level reconstruction.

\subsection{Data partitioning and conditioning controls}

The intended generalisation claim determines both the data partition and the independent inferential unit. Cross-patient claims require patient-level separation, whereas within-patient temporal transfer may be evaluated across sessions, seizures, or sufficiently separated recording blocks. Overlapping windows from the same recording are not independent observations and should be generated only after the underlying data have been partitioned. Data-dependent preprocessing, including normalisation, artefact rejection, contact selection, imputation, pretraining, and hyperparameter optimisation, should likewise be confined to the training partition. Temporal guard regions are required when filtering, windowing, or target-patient adaptation can transfer information across nominal split boundaries~\cite{Kapoor2023Leakage}.

The sampling hierarchy must also be reflected in statistical evaluation. Large numbers of windows from a small number of patients do not provide corresponding evidence of population-level generalisation. Patient-level claims therefore require aggregation and uncertainty estimates at the patient level, with lower-level observations treated as nested data. Reporting should include all eligible patients, contacts, and events, together with the denominator retained after any target-selection procedure. When performance is reported for a selected observable subset, corresponding all-target results or performance--coverage curves distinguish improved fidelity from reduced coverage.

Conditional validity requires comparison between explicitly defined information sets. Let \(B\) denote all non-concurrent information available at deployment, including anatomy, domain, task context, and any permissible target-patient iEEG adaptation data. A marginal predictor \(\boldsymbol{p}(\boldsymbol{y}\mid B)\) quantifies the structure recoverable without the concurrent scalp segment, whereas \(\boldsymbol{p}(\boldsymbol{y}\mid \boldsymbol{x},B)\) additionally conditions on scalp EEG. Simple linear mappings, scalp-only predictors, and ESI-derived estimates provide complementary baselines for architectural and non-invasive information content. Shuffled or time-shifted scalp--iEEG pairs, scalp ablation, anatomy-mismatched targets, task-label shuffling, and reference perturbations further test whether performance depends on the intended conditioning signal. A reproducible improvement from \(\boldsymbol{p}(\boldsymbol{y}\mid B)\) to \(\boldsymbol{p}(\boldsymbol{y}\mid \boldsymbol{x},B)\) after these controls provides direct evidence for an incremental contribution from concurrent scalp EEG.

\subsection{Claim-specific evaluation metrics}

Evaluation metrics should follow the estimand rather than the model architecture. For waveform reconstruction, Pearson correlation alone is insufficient because it is insensitive to gain and offset and can be dominated by shared low-frequency structure. Waveform evaluation should therefore combine correlation or concordance with amplitude and bias error, time-resolved morphology, and frequency-resolved fidelity. Spectral, phase, coherence, covariance, or connectivity metrics are informative only when the corresponding quantity forms part of the reconstruction claim. Event inference instead requires event-level sensitivity, PPV, false-alarm rate, precision--recall performance, latency, and, where relevant, morphology. Probabilistic outputs require proper scoring rules, predictive-interval coverage and width, stratified calibration, and risk--coverage analysis~\cite{Gneiting2007}.

These metrics quantify different evidential properties and should not be interpreted interchangeably. High spectral similarity does not establish correct event timing, and improved decoder performance can arise because synthetic or transferred representations regularise the downstream model rather than because the corresponding intracranial waveform was recovered. Conversely, a representation with limited waveform fidelity may preserve clinically or behaviourally relevant information. Evaluation against the specified target and intended use therefore separates waveform fidelity, event accuracy, representation quality, and downstream utility.

\subsection{Incremental utility and translational pathways}

Incremental utility asks whether virtual iEEG improves an intended decision beyond information already available from scalp EEG, ESI, or established workflows. In epilepsy, plausible near-term applications include event triage, prioritisation of candidate intervals, and support for implantation hypotheses~\cite{lam2016widespread,Lam2017,AbouJaoude2022,Zauli2024,Yamaguchi2026}. Relevant comparisons require tuned scalp-only and, where applicable, ESI or source-space workflows under identical data partitions. Blinded evaluation can then test changes in reference-standard concordance, review time, diagnostic yield, or other prespecified clinical endpoints at matched operating points.

For BCI applications, downstream decoder improvement represents a utility claim rather than evidence of contact-level reconstruction. Cross-modal pretraining and synthetic-representation studies have reported such gains~\cite{Dong2026Bridge,Yang2026CORTEG,Wang2026Synthetic}, but target-patient ECoG fine-tuning constitutes invasive adaptation rather than zero-shot non-invasive reconstruction. Incremental utility is therefore assessed against tuned EEG-only and source-space models under matched optimisation budgets, artefact controls, and session- or patient-transfer protocols~\cite{Blankertz2011,Lotte2018,Lawhern2018,Schirrmeister2017,Roy2019,jiang2026comprehensive,wang2026pa,zhao2026rmetnet,wang2026cfspmnet,zhao2025rgftslanet}.

In cognitive neuroscience, current evidence more directly supports hypothesis prioritisation than replacement of invasive measurements. Simultaneous studies can identify candidate intervals, frequency bands, or regions linking scalp and intracranial dynamics~\cite{Boran2019,Dimakopoulos2022,Pigorini2024}. Extrapolation from implanted epilepsy cohorts to different tasks, states, pathologies, or healthy populations changes several components of the deployment distribution simultaneously and therefore requires independent validation in the target domain.

Current reconstruction studies, including NeuroFlowNet, CAST, and Dong et al., have not demonstrated prospective effects on clinical decisions. Therefore, translational evidence remains downstream of measurement, conditional, and deployment validity. A progression from external-site or prospective silent-mode evaluation to blinded workflow studies and ultimately outcome assessment can test whether model outputs provide incremental value without being treated as direct measurements. Reporting frameworks such as TRIPOD+AI and DECIDE-AI provide relevant guidance for intended use, participant flow, missingness, uncertainty, human interaction, and failure handling~\cite{Collins2024TRIPODAI,Vasey2022DECIDEAI}.

\section{Outstanding questions and research priorities}

Current evidence raises several testable questions about the limits and utility of scalp-to-intracranial inference. Progress depends on identifying where concurrent scalp EEG adds information, where reconstruction remains unsupported, and whether virtual iEEG provides value beyond existing non-invasive measurements.

\begin{itemize}

\item \textit{How much information about a specified intracranial target is contributed by the concurrent scalp segment?}
Comparison of \(\boldsymbol{p}(\boldsymbol{y}\mid \boldsymbol{B})\) with \(\boldsymbol{p}(\boldsymbol{y}\mid \boldsymbol{x},\boldsymbol{B})\), together with target-reliability and reference-sensitivity estimates, can quantify the scalp-conditioned increment.

\item \textit{Where are the anatomical, spectral, state-dependent, and event-specific limits of recoverability?}
Performance should be stratified by region, depth, frequency, state, and event type, with stimulation and spontaneous recordings used as complementary rather than interchangeable evidence.

\item \textit{Can predictive uncertainty identify unsupported reconstructions?}
Calibration and risk--coverage analysis should determine whether uncertainty and abstention reliably identify high-error contacts, states, or anatomies.

\item \textit{What generalises without target-patient invasive data?}
Zero-shot claims require patient- and site-level evaluation without target-patient iEEG for fitting, calibration, contact selection, or hyperparameter choice.

\item \textit{Does virtual iEEG provide incremental utility beyond scalp EEG and ESI?}
Clinical, BCI and cognitive-neuroscience claims require endpoints that match their intended use and comparison with tuned non-invasive baselines.

\item \textit{Which negative results meaningfully define the reconstruction boundary?}
Well-powered failures under reliable targets and adequate controls can delineate anatomical, spectral, or state-dependent limits that pooled positive results may obscure.

\end{itemize}

Resolving these questions requires paired datasets with sufficient provenance to characterise anatomy, montage, synchronisation, hardware, state, event origin, target eligibility, and exclusions~\cite{Pigorini2024}. BIDS, EEG-BIDS, and iEEG-BIDS provide an established metadata foundation~\cite{Gorgolewski2016,Pernet2019,Holdgraf2019,Appelhoff2019}. Locked benchmarks should distinguish temporal, patient, and site transfer and explicitly separate adaptation from target-iEEG-free evaluation. Fixed splits, common baselines, patient-level reporting, anatomy- and frequency-stratified performance, target coverage, and overlap audits would make cross-study comparisons more interpretable. A cumulative evidence base should value boundary results alongside positive reconstruction scores. Recoverability within a defined operating regime is scientifically informative, whereas increasingly realistic generated waveforms provide limited evidence of progress if conditional dependence, calibration, coverage, and incremental utility remain unresolved.

\section{Conclusion}
This review establishes a target-centred framework for evaluating scalp-to-intracranial reconstruction (STIR) and virtual iEEG, distinguishing them from EEG source imaging (ESI): ESI estimates neural generators in source space, whereas STIR predicts iEEG-defined events, features, representations, or contact-level waveforms from scalp EEG, with outputs assigned intracranial semantics termed virtual iEEG. Existing studies demonstrate that scalp EEG can support inference of selected intracranial events and representations, as well as waveform reconstruction under paired, patient-specific, or adapted cross-patient settings. However, current evidence remains target- and regime-dependent and does not establish unique recovery of arbitrary contact-level activity or general target-iEEG-free deployment. Progress therefore requires clear separation of observability, predictability, identifiability, fidelity, and utility, together with conditioning controls, matched scalp-EEG and ESI baselines, calibrated uncertainty, independent patient- and site-level validation, and prospective utility testing. Such evidence will define when virtual iEEG provides reliable incremental information beyond existing non-invasive approaches.

\section*{Acknowledgments}
This work was supported by  The Hong Kong Polytechnic University Departmental Collaborative Research Fund (Project ID: P0056428), an internal grant from The Hong Kong Polytechnic University (Project ID: P0048377),The Hong Kong Polytechnic University Collaborative Research with World-leading Research Groups Fund (Project ID: P0058097), and Research Grants Council Collaborative Research Fund (Ref: C5033-24G).

\bibliographystyle{unsrt}  
\bibliography{references}  

\end{document}